\documentclass[letterpaper]{article} % DO NOT CHANGE THIS
\usepackage{aaai23}  % DO NOT CHANGE THIS
\usepackage{times}  % DO NOT CHANGE THIS
\usepackage{helvet}  % DO NOT CHANGE THIS
\usepackage{courier}  % DO NOT CHANGE THIS
\usepackage[hyphens]{url}  % DO NOT CHANGE THIS
\usepackage{graphicx} % DO NOT CHANGE THIS
\usepackage{natbib}  % DO NOT CHANGE THIS AND DO NOT ADD ANY OPTIONS TO IT
\usepackage{caption} % DO NOT CHANGE THIS AND DO NOT ADD ANY OPTIONS TO IT
\usepackage{makecell}
\usepackage{xcolor}
\usepackage{soul}

\usepackage{algorithm}
\usepackage{algorithmic}

\usepackage{newfloat}
\usepackage{listings}
\DeclareCaptionStyle{ruled}{labelfont=normalfont,labelsep=colon,strut=off} % DO NOT CHANGE THIS
\floatstyle{ruled}
\newfloat{listing}{tb}{lst}{}
\floatname{listing}{Listing}
\title{Grape Cold Hardiness Prediction via Multi-Task Learning}
\author{
    Aseem Saxena\textsuperscript{\rm 1}, Paola Pesantez-Cabrera\textsuperscript{\rm 2}, Rohan Ballapragada\textsuperscript{\rm 1}, Kin-Ho Lam\textsuperscript{\rm 1}, Markus Keller\textsuperscript{\rm 3} and Alan Fern\textsuperscript{\rm 1}
}
\affiliations{
    \textsuperscript{\rm 1}School of Electrical Engineering and Computer Science, Oregon State University\\
    \textsuperscript{\rm 2} School of Electrical Engineering and Computer Science, Washington State University\\
    \textsuperscript{\rm 3}Irrigated Agriculture Research and Extension Center, Washington State University\\ 
    saxenaa@oregonstate.edu, p.pesantezcabrera@wsu.edu, ballaprr@oregonstate.edu, lamki@oregonstate.edu, mkeller@wsu.edu, alan.fern@oregonstate.edu 
    
}

\usepackage{bibentry}
\begin{document}

\maketitle

\begin{abstract}
Cold temperatures during fall and spring have the potential to cause frost damage to grapevines and other fruit plants, which can significantly decrease harvest yields. To help prevent these losses, farmers deploy expensive frost mitigation measures such as sprinklers, heaters, and wind machines when they judge that damage may occur. This judgment, however, is challenging because the cold hardiness of plants changes throughout the dormancy period and it is difficult to directly measure. This has led scientists to develop cold hardiness prediction models that can be tuned to different grape cultivars based on laborious field measurement data. In this paper, we study whether deep-learning models can improve cold hardiness prediction for grapes based on data that has been collected over a 30-year time period. A key challenge is that the amount of data per cultivar is highly variable, with some cultivars having only a small amount. For this purpose, we investigate the use of multi-task learning to leverage data across cultivars in order to improve prediction performance for individual cultivars. We evaluate a number of multi-task learning approaches and show that the highest performing approach is able to significantly improve over learning for single cultivars and outperforms the current state-of-the-art scientific model for most cultivars.
\end{abstract}

\section*{Introduction}
% Grapevines have the ability to survive cold temperatures during fall, winter, and spring, this is known as Cold Hardiness ($H_c$). Cold hardiness is dynamic in nature with a predictable seasonal trend; cold hardiness is low at the beginning of fall as the plant has not yet acclimatized and peaks during mid-winter when the plant reaches acclimation, reducing with the onset of spring as the plant deacclimatizes. Acclimation in grapevines happens by shedding leaves and reaching dormancy. Depending on the plant's current cold hardiness, cold temperatures can turn out to be potentially lethal, especially sudden frost events. Several active and expensive preemptive methods like wind machines, sprinklers, and heaters can be deployed to mitigate this. Since Cold Hardiness is dynamic, growers can't simply use a single threshold temperature for frost mitigation. Also, deploying frost mitigation at all times is not an economically viable solution. The decision is further complicated by the fact that cold hardiness varies for different grape varieties. 

The ability of grapevines to survive cold temperatures during fall, winter, and spring, is known as Cold Hardiness ($H_c$). Cold hardiness in grapes and other plants is dynamic in nature with a predictable seasonal trend. Cold hardiness is low at the beginning of fall as the plant has not yet acclimatized and peaks during mid-winter when the plant reaches acclimation. As spring arrives, the plant deacclimatizes and the cold hardiness decreases to the low summer levels. This means that during the fall and spring, when cold hardiness is low, unusually cold temperatures can be lethal, especially arising from sudden frost events. 

To mitigate lethal damage due to cold temperatures, farmers can deploy expensive preemptive methods such as wind machines, sprinklers, and heaters to raise the air temperature. However, the decision of when to invest in expensive mitigation depends on knowledge of the current unknown cold hardiness. While cold hardiness can be measured, it requires expertise and expensive equipment, which farmers rarely have. Thus, farmers rely on estimates of cold hardiness derived from a combination of experience and scientific models. This highlights the need for accurate data-centric models for cold-hardiness prediction. 

Current state-of-the-art cold hardiness models (e.g. \cite{ferguson_modeling_2014}) use a biological basis to obtain a parameterized model that can be tuned for different grape cultivars using cold-hardiness data. While reasonably effective, these models are relatively simple and only use ambient temperature as input. Rather, cold hardiness likely depends on multiple weather factors (e.g. humidity and precipitation) in complex ways \cite{mills_cold-hardiness_2006} that are not full captured by current scientific models. This raises the question of whether modern machine learning methods can improve on current models via their increased expressiveness and ability to consume richer inputs.  

% This highlights the need for accurate data-centric cold hardiness prediction models which growers can use for better decision-making. With this motivation, Ferguson et al. developed a discrete dynamic model of grapevine cold hardiness which applies a biological basis to parameterize the model \cite{ferguson_modeling_2014}. The model is based on domain understanding, is easy to interpret, and works sufficiently well for most cultivars. The model suffers from a lack of expressiveness and relies only on ambient temperature as an explanatory variable, even though surface moisture has been shown to affect cold hardiness as well \hl{(citation to paper that uses or states surface moisture is important)}. 

%In this work, we tackle the limitations of current models by studying Cold hardiness prediction as a time series problem and exploit the flexibility and expressivity of Recurrent Neural Networks to model it.  \\
%Neural Networks suffer from overfitting when trained with insufficient amounts of data. We mitigate that by building multi-task models which efficiently use the data of all cultivars and can be adapted to predict cold hardiness for any cultivar.  

In this work, we evaluate the use of Recurrent Neural Networks (RNNs) for predicting cold hardiness based on time series weather data. A key challenge is that ground-truth cold-hardiness data is quite limited in comparison with many applications of deep learning. In our experiments, we find that for some grape cultivars, where there is significant data, RNNs can be quite accurate and outperform a current state-of-the-art model. However, for cultivars with more limited data, the RNNs can perform poorly. This raises the question of whether we can leverage data across multiple cultivars to improve the prediction performance for cultivars with limited data. 

Our main contributions are: 1) To frame this multi-cultivar learning problem as multi-task learning, and 2) To propose and evaluate a variety of multi-task RNN models on real-world data collected from over twenty cultivars with data amounts ranging from 34 to just 4 seasons. Our results show that multi-task learning is able to significantly outperform learning from just the data of a single cultivar and very often outperforms the state-of-the-art scientific model. We are aiming to install a model result from this work on an existing weather network for trial use by grape farmers.

\begin{figure}[t]
\centering
\includegraphics[width=1\columnwidth]{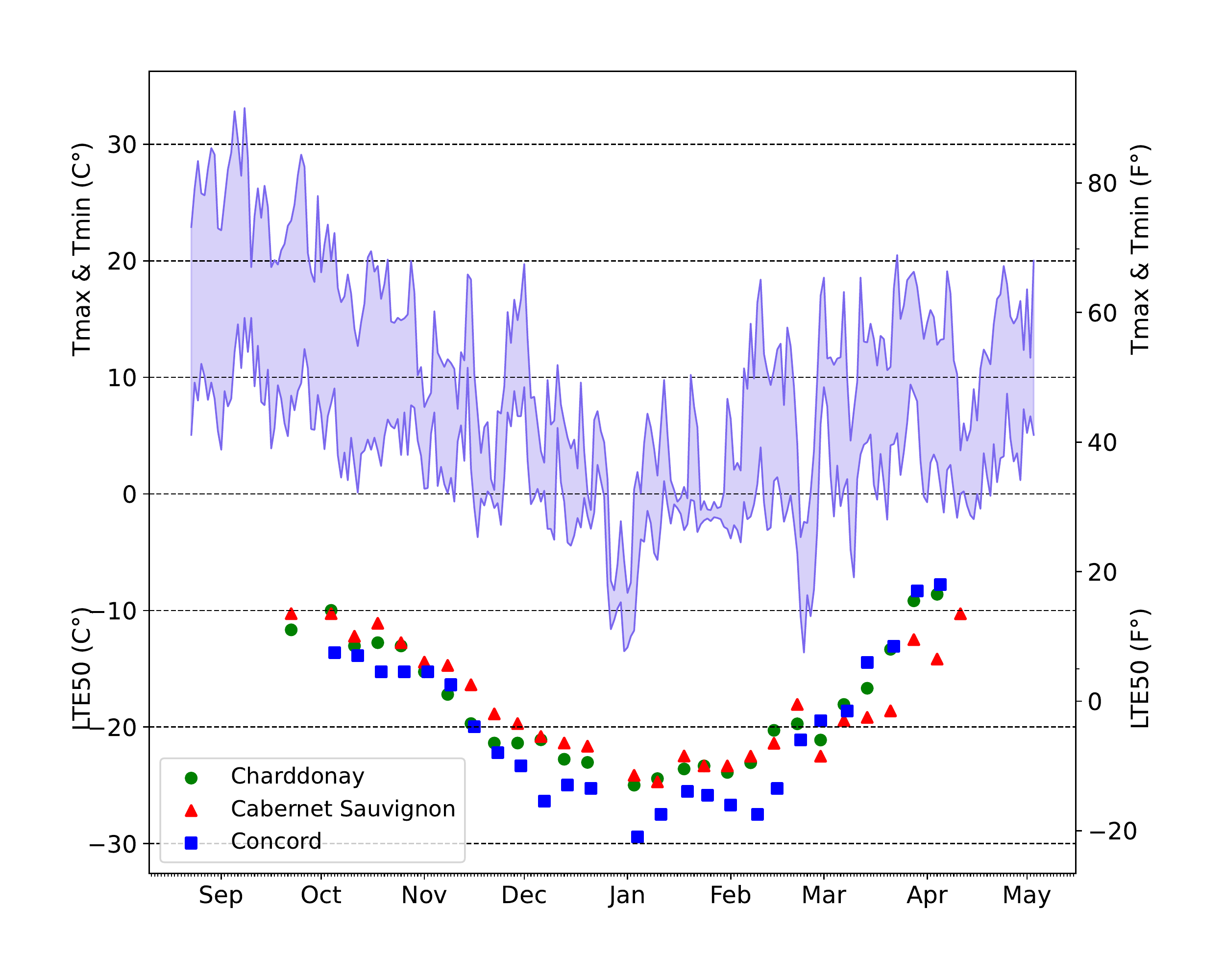} % Reduce the figure size so that it is slightly narrower than the column. Don't use precise values for figure width.This setup will avoid overfull boxes.
\caption{$LTE_{50}$ measurements for three cultivars and temperature range for one season. The bowl shape indicates an initial increase in cold hardiness (decrease in lethal temperature) during acclimation followed by a decrease in cold hardiness during deacclimation. Notice the difference in variation of cold hardiness for the chosen cultivars.}
\label{fig:LTE-example-seasons}
\end{figure}

\section*{Background}

The \emph{cold hardiness} of a plant characterizes its ability to resist injury during exposure to low temperatures. In this work, we will focus on grapevine cold hardiness, where injury corresponds to lethal bud freezing, which decreases crop yield. In order to quantify how grapevine cold hardiness varies throughout the dormancy period, scientists use differential thermal analysis (DTA). DTA results in a measurement of the lethal temperatures at which 10\%, 50\%, and 90\% of the bud population die/freeze, which are denoted by $LTE_{10}$, $LTE_{50}$, and $LTE_{90}$ respectively \cite{mills_cold-hardiness_2006}. Figure \ref{fig:LTE-example-seasons} shows the $LTE_{50}$ values for three cultivars and temperature ranges throughout a dormant season. 
 %This involves putting bud samples in a thermoelectric module, which can sense low temperature exotherms (LTEs) resulting from the freezing of individual buds. The module is placed in a controlled refrigerator, which systematically decreases the temperature as LTEs are monitored. 

Since this measurement process requires expensive, specialized equipment and expertise, scientists have used collected data to develop grape cold-hardiness models, which aim to estimate the lethal temperatures based on only historical temperature data. The current state-of-the-art model, developed by \citeauthor{ferguson_dynamic_2011} (\citeyear{ferguson_dynamic_2011}), integrates plant biology concepts to find a relation between daily temperatures and changes in cold hardiness. Intuitively, the \emph{Ferguson model} computes the daily change in cold hardiness (e.g. as measured by $LTE_{50}$) based on the day's accumulated thermal time (being above or below certain temperature thresholds) weighted by coefficients that vary with the stage of dormancy. The model has a small number of parameters, e.g. thresholds for thermal times, which can be tuned for a particular cultivar. Tuning was done by performing a brute-force grid search over the parameter space to identify the parameter settings that resulted in the most accurate predictions. While this model has produced promising results and is in use by growers, it has limited expressiveness (only a handful of parameters) and only uses daily temperature data as input, rather than also factoring in other influential weather measurements (e.g. humidity and precipitation). 

The limitations of the current scientific models raise the question of whether we can improve cold-hardiness prediction through the use of modern deep learning models. On one hand, such black-box deep models can be much more expressive and can easily incorporate additional weather data as input. On the other hand, the cold-hardiness data set sizes are relatively small from a deep-learning perspective, which may limit the potential benefits. The remainder of the paper explores this question. Below we first describe the cold-hardiness data sets used in our work followed by a description and evaluation of our deep learning approaches.  

\section*{Cold Hardiness Datasets}

The cold hardiness of endo\textendash and ecodormant primary buds from up to 30 genetically diverse cultivars/genotypes of field-grown grapevines has been measured since 1988 in the laboratory of the WSU Irrigated Agriculture Research and Extension Center (IAREC) in Prosser, WA (46.29°N latitute; -119.74°W longitude). In the vineyards of the IAREC, the WSU-Roza Research Farm, Prosser, WA (46.25°N latitude; -119.73°W longitude), and in the cultivar collection of Ste. Michelle Wine Estates, Paterson, WA (45.96°N latitute; -119.61°W longitude), cane samples containing dormant buds were collected daily, weekly, or at 2-week intervals from leaf fall in autumn to bud swell in spring. These two phenological events typically occurred in October and in April, respectively \cite{ferguson_dynamic_2011, ferguson_modeling_2014}. 

% (generally at the time of the first frost)
%Then, those dormant buds were analyzed following the protocol described by \cite{mills_cold-hardiness_2006}, using Differential Thermal Analysis (DTA) that measures low-temperature exotherms (LTE). An LTE corresponds to the \emph{lethal} temperature at which a percentage of the primary buds die. Thus, cold hardiness could then be expressed as $LTE_{50}$, which is the lethal temperature for 50\% of buds tested. 
%
All samples were analyzed with DTA to record ground truth for $LTE_{10}, LTE_{50}$, and $LTE_{90}$ measurements of cold hardiness.  
Additionally, meteorological/environmental daily data from the closest on-site weather station to each vineyard (cultivar) was obtained using the API provided by AgWeatherNet \cite{AgWeatherNet}. The three stations used are Prosser.NE (46.25°N latitude; -119.74°W longitude), Roza.2 (46.25°N latitude; -119.73°W longitude), and Paterson.E (45.94°N latitude; -119.49°W longitude).

The result is a continually growing dataset for each cultivar that contains a varying number of seasons of daily weather data along with cold-hardiness LTE labels for the days that samples were collected. Following prior work we consider \emph{a season} to extend from September 7th to May 15th, which is a conservative interval that should almost always contain the full dormancy period. Our experiments involve cultivars with data sets ranging from 34 to 4 seasons. 

%{\bf Cultivars Data Summary.}
\begin{table}[htp]
\centering
\fontsize{10}{10}\selectfont
\resizebox{1\columnwidth}{!}{
\begin{tabular}{ |l|l|r|r| }
\hline
\multicolumn{1}{|c|}{\textbf{Cultivar}}             &\makecell{\textbf{LTE} \\ \textbf{Data Seasons}}                      &\makecell{\textbf{LTE Total} \\ \textbf{Years of Data}}    &\makecell{\textbf{LTE Total} \\ \textbf{Samples}} \\\hline
%\thead{Cultivar}     &\thead{LTE Data Seasons}                      &\thead{LTE Total}    &\thead{LTE Total}\\
%                     &                                              &\thead{Years of Data}            &\thead{Samples}\\\hline
Barbera	             &2006-2022	                        &14                         &151\\\hline
Cabernet Franc	     &2005-2012	                        &4                          &35\\\hline
Cabernet Sauvignon	 &1988-2022	                                    &34                         &829\\\hline
Chardonnay	         &1996-2022	                                    &26                         &783\\\hline
Chenin Blanc	     &1988-2022	            &18                         &193\\\hline
Concord	             &1988-2022	&27                         &484\\\hline
Gewurztraminer	     &2005-2016	                        &9                          &101\\\hline
Grenache	         &2006-2022	                        &14                         &151\\\hline
Lemberger	         &2006-2016	                        &6                          &60\\\hline
Malbec	             &2004-2022	                        &17                         &261\\\hline
Merlot	             &1996-2022	                                    &26                         &897\\\hline
Mourvedre	         &2005-2022	                        &12                         &133\\\hline
Nebbiolo	         &2006-2022	                        &14                         &152\\\hline
Pinot Gris	         &2003-2022	                        &17                         &190\\\hline
Riesling	         &1988-2022	                                    &34                         &636\\\hline
Sangiovese	         &2005-2022                        &15                         &165\\\hline
Sauvignon Blanc	     &2006-2022	            &12                         &140\\\hline
Semillon	         &2006-2022	                        &13                         &201\\\hline
Syrah	             &1999-2022	                                    &23                         &486\\\hline
Viognier	         &1999-2022	                        &18                         &206\\\hline
Zinfandel	         &2006-2022	                        &14                         &150\\\hline
\end{tabular}
}
\caption{Summary of cultivars' LTE data collection.} %There were several seasons for which there was no LTE data collected.}
\label{tab:data-description}
\end{table}

{\bf Cultivar Dataset Details.}
%The complete data sets for each cultivar are available at \url{https://github.com/...removed_for_anonymous_review...}.% 
Table \ref{tab:data-description} shows a summary of the number of years of data collected for selected cultivars. 
The dataset for a given cultivar contains a row for each day of all data-collection seasons. Note, since cold hardiness was not measured on each day of a season, some rows  do not contain LTE data. Below we highlight the key information contained in each row used by our models. 
%\url{https://github.com/AgAIDInstitute/Public-Data/tree/main/ColdHardiness/Grapes/Cultivars_LTE}
\begin{itemize} 
 \item DATE: The date of the weather observation. 
 \item AWN\_STATION: The closest AgWeatherNet station from where the environmental readings are taken.
% \item SEASON: A dormant season usually goes from October to mid-April. However, following the instructions to run the current cold hardiness model provided by WSU, a season is considered from September 7th (JDAY 250) to May 15th (JDAY 500).
% \item SEASON\_JDAY: Julian day. It is the integer that represents the count of the number of days that have passed from the beginning of a period. For a dormant season, the JDAY continues at the new year (i.e., Dec 31 is JDAY 365, then the next day Jan 1 is JDAY 366, and so on).
%\item DORMANT\_SEASON: It will have a value of 1 if the date belongs to the dormant season or 0 if it does not.
% \item LTE10 (when available): Lethal temperature observed from vineyard samples when 10\% of the buds die. In degrees Celsius. 
% \item LTE50 (when available): Lethal temperature observed from vineyard samples when 50\% of the buds die. In degrees Celsius.
% \item LTE90 (when available): Lethal temperature observed from vineyard samples when 90\% of the buds die. In degrees Celsius.
 \item LTE values (when available): $LTE_{10}$, $LTE_{50}$, $LTE_{90}$. In degrees Celsius. 
% \item LTE50 (when available): In degrees Celsius.
% \item LTE90 (when available): In degrees Celsius.
%\item YEAR\_JDAY: Julian day of the given date.
 \item MIN\_AT, AVG\_AT, MAX\_AT: Minimum, average, and maximum air temperature observed at 1.5 meters above the ground. In degrees Celsius. 
% \item MIN\_AT, AVG\_AT, MAX\_AT: Minimum, average, and maximum air temperature observed at 1.5 meters above the ground if the temperature sensor is installed; otherwise, the value will be NaN. If the AgWeatherNet station is missing the readings, the value will be -100. In degrees Celsius. 
\item MEAN\_AT: $(MIN\_AT + MAX\_AT)/2$. In degrees Celsius.\footnote{This is recorded since it is the temperature measure used by the scientific model.}
 \item MIN\_RH, AVG\_RH, MAX\_RH: Minimum, average, and maximum relative humidity value observed at 1.5 meters above the ground. In percent.
 \item MIN\_DEWPT, AVG\_DEWPT, MAX\_DEWPT: Minimum, average, and maximum dew point (temperature the air needs to be cooled to in order to achieve relative humidity). In degrees Celsius.
 \item P\_INCHES: Observed sum of precipitation for the daily period. In inches.
 \item WS\_MPH, MAX\_WS\_MPH: Average and maximum observed wind speed at 1.5 meters above the ground for the daily period. In Miles Per Hour. 
\end{itemize}

\section*{Deep Cold-Hardiness Models and Training}

\begin{figure*}[t]
\centering
\includegraphics[width=1\textwidth]{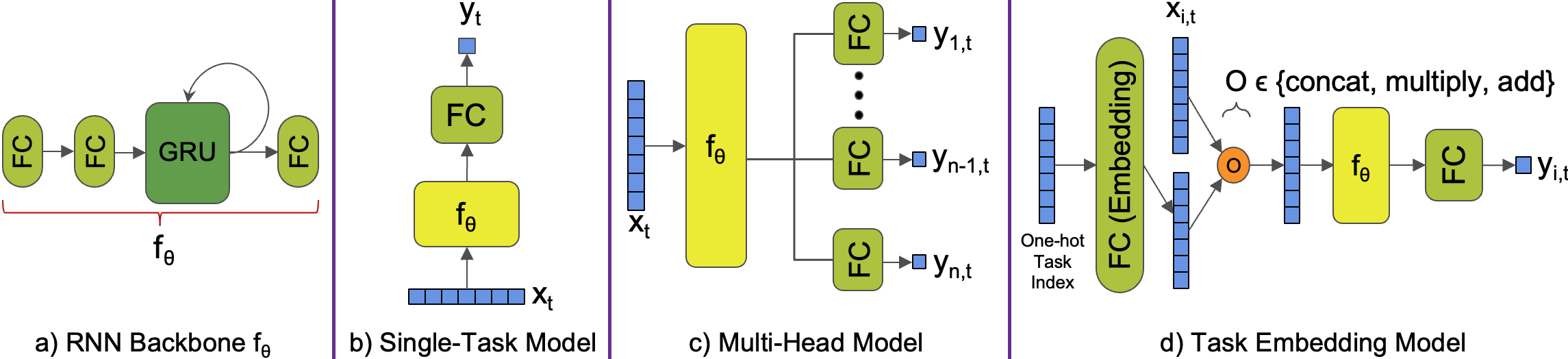} % Reduce the figure size so that it is slightly narrower than the column. Don't use precise values for figure width.This setup will avoid overfull boxes.
\caption{Network Architectures. FC denotes fully connected layers and GRU denotes Gated Recurrent Unit. a) The RNN backbone is used to process weather data sequences $(x_t)$. b) The single-task model with a single prediction head for a single cultivar. c) Multi-Head Model which has a prediction head for each cultivar allowing backbone features to be shared. d) Task Embedding Model, which combines the weather data features with a learned task embedding for each cultivar before entering the backbone network. }
\label{fig:model-diagrams}
\end{figure*}

Given the availability of cold-hardiness data, we can formulate cold-hardiness prediction as a sequence prediction problem. We will use $i$ to index the different grape cultivars with $N_i$ denoting the number of seasons available for cultivar $i$. The sequence data for season $k$ of cultivar $i$ is denoted by $S_{i,k}$ and has the form $S_{i,k} = (x_1, y_1, x_2, y_2, \ldots, x_H, y_H)$, where $x_t$ is the weather data for day $t$, $y_t$ is the ground truth LTE data for day $t$, and $H$ is the number of days per season. Recall that $y_t$ is not measured on each day of a season (e.g. measured every two weeks) and hence for days where the LTE measurements are unavailable $y_t = N/A$. Finally, the data set for cultivar $i$ is denoted by $D_i = \{S_{i,k} \;|\; k\in \{1,\ldots,N_i\}\}$.  

Given a data set $D_i$ our learning goal is to produce a model $M_i$ that can take as input a sequence of daily weather measurements $(x_1,x_2,\ldots, x_t)$ up to a particular day $t$ and produce a sequence of predicted LTE estimates $(\hat{y}_1,\hat{y}_2,\ldots,\hat{y}_t)$ for cultivar $i$. Typically, a farm manager will be most interested in the estimate $\hat{y}_t$. This estimate can then be compared to the low-temperature forecast for that day to help decide whether to prepare for frost mitigation measures. The key question of this work is to evaluate whether modern deep learning methods can provide farm managers with improved predictions compared to the current state-of-the-art cold-hardiness models. 

We will refer to the problem of learning $M_i$ based on only $D_i$ as \emph{single-task learning (STL)}, which is the general framework used for the vast majority of deep learning applications. Importantly, the performance of STL is significantly influenced by the amount of available training data, which according to Table \ref{tab:data-description} varies widely across the different cultivars. Thus, we might expect STL performance for cultivars with small datasets to suffer in comparison to those with large datasets. To address this issue, we consider the \emph{multi-task learning (MTL)} framework \cite{mtlfirstpaper, mtlsurvey1, mtlsurvey2,mtlsurvey3}, which involves learning a predictive model for cultivar $i$ using a combined dataset of all cultivars $D = \{D_1, D_2, \ldots, D_C\}$, where $C$ is the number of cultivars for which we have data. Intuitively, MTL offers the potential to identify common structures among the multiple learning tasks (i.e. cultivars) in order to improve performance for individual cultivars, especially those with limited data.  

Below, we first introduce the STL deep model that we developed, which will serve as our deep-learning baseline for cold-hardiness prediction. Next, we introduce two frameworks for modifying that model to support MTL. Finally, we describe certain details of the training strategy used in our experiments. To the best of our knowledge, this is the first work that has considered deep models for cold-hardiness prediction in both the STL and MTL settings.

\subsection*{Single-Task Model}

% The basic single-task model must processes sequences of weather data and output an LTE prediction for each day based on the current and past weather data. In this work, we use recurrent neural networks (RNNs) \cite{RNN} to process the sequence and form an internal state that represents the weather history at each day. Figure \ref{}a shows the backbone that is used in all of our models, denoted by $f_{\theta}$ with parameters $\theta$. The backbone network contains two fully connected (FC) layers, followed by a gated recurrent unit (GRU) layer\cite{GRU}, which is followed by another FC layer. Our single-task model simply adds another FC layer to this backbone and trains the parameters separately for each cultivar. The result is a set of learned parameters for each cultivar, which are used at test time for prediction. 

Our basic STL makes causal LTE predictions by sequentially processing a weather data sequence $x_1,x_2,\ldots, x_t$ and at each step outputting the corresponding LTE estimate.  For this purpose we use a recurrent neural network (RNN) model \cite{RNN}, which is a widely used model for sequence data. The RNN backbone used by both our STL and MTL models is illustrated in Figure \ref{fig:model-diagrams}a), which we denote by $f_{\theta}$ with parameters $\theta$. The backbone network begins with two Fully Connected (FC) layers, followed by a Gated Recurrent Unit (GRU) layer \cite{GRU}, which is followed by another FC layer. Our STL model, shown in Figure \ref{fig:model-diagrams}b), simply feeds daily weather data $x_t$ into the first FC layer as input and adds an additional FC layer to produce the final LTE prediction output. Intuitively, the GRU unit, through its recurrent connection is able to build a latent-state representation of the sequence data that has been processed so far. For our cold-hardiness problem, this representation should capture information about the weather history which is useful for predicting LTE. In some sense, the latent state can be thought of as implicitly approximating the internal state of the plant as it evolves during dormancy. As described below, each STL model $M_i$ is trained independently on its cultivar-specific dataset $D_i$.

%Each $T_i$ has its own $f_{\theta_i}$ with no parameters being shared across ttasks.  $L = L_i(g_{\phi_i}(f_{\theta{_i}}(x_{i_t})),y_{i_t})))$ with each loss being optimized separately. 

\subsection*{Multi-Task Models}

We consider two types of MTL models that directly extend the RNN backbone of Figure \ref{fig:model-diagrams}a), the multi-head model and the task embedding model.

{\bf Multi-Head Model.} The multi-head model is perhaps the most straightforward approach to MTL and has been quite successful in prior work when tasks are highly related \cite{mtlfirstpaper}. As illustrated in Figure \ref{fig:model-diagrams}c), the multi-head model is identical to the STL model, except, that it adds $C$ parallel cultivar-specific fully-connected layers to the backbone (i.e. prediction heads). Each prediction head is responsible for producing the LTE prediction for its designated cultivar. This model allows the cultivars to share the features produced by the RNN backbone, with each cultivar-specific output simply being a linear combination of the shared features. Intuitively, if there are common underlying features that are useful across cultivars, then this architecture allows those to emerge based on the combined set of data. Thus, cultivars with small amounts of data can leverage those useful features and simply need to tune a set of linear weights based on the available data. We abbreviate this model with \textbf{MultiH} in future sections.

% Depending on the cultivar, only the output of that cultivar's layer is computed and its loss is back-propagated. We tried weighting the losses for the MultiHead approach by the inverse of the task frequency (proportion of task-specific data in the aggregated dataset), which didn't lead to considerable gains in performance. We abbreviate the model with MultE in the following sections.

% The MultiHead Model starts with the backbone network and adds multiple cultivar-specific layers, all connected to the final layer of the backbone network. \cite{mtlfirstpaper} Depending on the cultivar, only the output of that cultivar's layer is computed and its loss is back-propagated. We tried weighting the losses for the MultiHead approach by the inverse of the task frequency (proportion of task-specific data in the aggregated dataset), which didn't lead to considerable gains in performance. We abbreviate the model with MultE in the following sections.

%$f_\theta$ is shared among all $T_i \epsilon T$.  $L = \sum L_i(g_{\phi_i}(f_\theta(x_{i_t})),y_{i_t}))))$.  
\begin{table*}[tb]
\centering\fontsize{9}{9}\selectfont
\resizebox{1.33\columnwidth}{!}{
\begin{tabular}{|l|c|c|c|c|c|c|}
\hline
\thead{Cultivar} & \thead{MultE} & \thead{ConcatE} & \thead{AddE} & \thead{MultiH} & \thead{Single} & \thead{Ferguson} \\ \hline
Barbera            & 1.92 & 1.50 & 2.07 & 1.89 & 4.22 & 1.78 \\ \hline
Cabernet Franc     & 4.84 & 2.36 & 3.49 & 2.39 & 4.00 & 1.45 \\ \hline
Cabernet Sauvignon & 2.93 & 1.75 & 1.82 & 2.27 & 3.43 & 1.83 \\ \hline
Chardonnay         & 1.33 & 1.46 & 1.44 & 1.40 & 1.60 & 1.79 \\ \hline
Chenin Blanc       & 1.85 & 1.51 & 1.57 & 1.45 & 2.47 & 2.27 \\ \hline
Concord            & 2.33 & 2.42 & 2.32 & 1.98 & 2.61 & 2.02 \\ \hline
Gewurztraminer     & 1.97 & 1.40 & 1.66 & 1.20 & 2.70 & 1.84 \\ \hline
Grenache           & 3.07 & 1.86 & 2.17 & 1.79 & 2.86 & 1.92 \\ \hline
Lemberger          & 3.01 & 1.65 & 2.24 & 1.49 & 3.23 & 2.21 \\ \hline
Malbec             & 1.80 & 1.32 & 1.32 & 0.96 & 1.71 & 1.66 \\ \hline
Merlot             & 1.74 & 1.53 & 1.39 & 1.53 & 1.66 & 1.55 \\ \hline
Mourvedre          & 1.84 & 1.65 & 1.70 & 1.56 & 2.25 & 1.83 \\ \hline
Nebbiolo           & 2.36 & 1.58 & 1.87 & 1.24 & 2.48 & 1.80 \\ \hline
Pinot Gris         & 2.07 & 1.61 & 1.63 & 1.61 & 2.04 & 2.02 \\ \hline
Riesling           & 2.80 & 1.47 & 1.77 & 1.97 & 3.63 & 1.55 \\ \hline
Sangiovese         & 1.65 & 1.73 & 1.71 & 1.40 & 1.84 & 1.61 \\ \hline
Sauvignon Blanc    & 1.33 & 1.43 & 1.52 & 1.22 & 1.71 & 1.42 \\ \hline
Semillon           & 2.37 & 1.67 & 1.42 & 1.75 & 3.58 & 1.50 \\ \hline
Syrah              & 1.22 & 1.22 & 1.28 & 1.29 & 1.57 & 1.25 \\ \hline
Viognier           & 3.90 & 1.75 & 2.30 & 2.28 & 4.16 & 1.36 \\ \hline
Zinfandel          & 3.10 & 1.45 & 1.56 & 1.60 & 2.64 & 1.90 \\ \hline
\end{tabular}%
}
\caption{Comparison of the performance of proposed MTL methods with STL and the existing state-of-the-art method. Note that the performance is measured in terms of Root Mean Squared Error.}
\label{tab:mainresults}
\end{table*}
{\bf Task Embedding Models.} Our proposed Task Embedding models are motivated by thinking about current scientific models and how they address multiple tasks. The Ferguson model, for example, has a fixed structure, based on scientific knowledge, but a small number of parameters that can be tuned for each cultivar. Our task embedding model aims to generalize this concept by having a neural network learn both the structure of the model that accepts task-specific parameters as well as learning the parameters of each cultivar. Note that the cultivar parameters and model structure will not have a clear scientific interpretation due to the black-box nature of deep models. The trade-off for interpretability is the potential for better performance due to increased expressive power. 

Specifically, our proposed Task Embedding models, as shown in Figure \ref{fig:model-diagrams}d), are similar in spirit to context-sensitive neural networks \cite{csnn,taskembedding1,taskembedding2}, where a task-specific context is provided as additional input to the neural network with only a single output being computed. 

We obtain this task-specific context by encoding the task as a one-hot vector and finding a corresponding mapping using a differentiable embedding layer. We explore different ways of incorporating the obtained task embedding, via element-wise Addition, Concatenation, and element-wise Multiplication. We abbreviate these models with \textbf{AddE}, \textbf{ConcatE}, \textbf{MultE} in future sections.

\subsection*{Model and Training Details}
% Things to include: 1) We train with MSE loss. Our setting is somewhat unusual in that the ground truth is only available for some of the time steps in the sequences. Thus we mask the loss and only accumulate loss at timesteps with ground truth. 2) In all experiments our network architectures use the following dimenions: .... 3) The input weather features in all experiments are ... 4) other details

We construct our dataset by selecting the dormant season data for all cultivars. Missing features are filled in by linear interpolation. We discard seasons that have $ <10\% $ valid LTE readings. We only include seasons where at least $ 90\% $ of temperature data is not missing. Missing LTE label readings are not interpolated, instead, the missing LTE labels' losses are masked during the training and evaluation process. We choose 2 seasons for each cultivar as our test set. We run three trials of training for all our experiments with different train/test splits and average the performance over the three trials.

We rely on the following weather features for learning our models - Temperature, Humidity, Dew Point, Precipitation, and Wind Speed. Our models output predictions for $LTE_{10}$, $LTE_{50}$, and $LTE_{90}$ which are optimized simultaneously, helping in inductive transfer. We focus exclusively on the model's predictive power for $LTE_{50}$ in this work. We consider the Mean Squared Error(MSE) as our loss function and treat the Root Mean Squared Error(RMSE) as our performance metric. 
We consider Adam \cite{adam} as the optimizer of choice for our training process. We use a learning rate of 0.001 with a batch size of 12 seasons shuffled randomly. We train all our models for 400 epochs. The input features have a dimensionality of 12. The output dimensionality of the linear layers of the RNN backbone are 1024, 2048 and 1024 respectively. The GRU has a hidden state and internal memory of dimensionality 2048.
\section*{Experiments}

In this section, we present our main empirical results. Our experiments involve 21 cultivars from Table \ref{tab:data-description}. In particular, we removed any cultivar that has less than 4 years of data and removed cultivars for which the Ferguson model results were unavailable for comparison. 

{\bf Multi-Task Versus Single-Task Learning.} Table \ref{tab:mainresults} shows the root mean squared error (RMSE) of the multi-task models, single-task model, and Ferguson model for all 21 cultivars. The first observation is that with the exception of Chardonnay, the single-task model never outperforms the state-of-the-art Ferguson model. For cultivars with small amounts of data, there is often a dramatic decrease in performance over Ferguson, while other cultivars with larger datasets are close to Ferguson's performance. 

The second observation is that for each cultivar, with rare exceptions, the 4 multi-task models all outperform the corresponding single-task model. As expected, the improvement tends to be most pronounced for the smaller dataset cultivars. This shows that multitask learning is indeed able to identify and exploit common structures among the different cultivars, leading to improved generalization. Among the MTL methods, the MultiHead approach consistently outperforms other methods. Among the task embedding approaches, the concatenation approach performs best. 

We observe, with the exception of Cabernet Franc and Viognier, that our approaches outperform the Ferguson model, the MultiHead or concat task embedding approaches being the best-performing models for most cultivars. The gap in the performance is more dramatic in cultivars with low data, such as  Lemberger and Gewurztraminer.

\begin{table}[t]
\centering \fontsize{9}{9}\selectfont
\resizebox{1\columnwidth}{!}{
\begin{tabular}{|l|c|c|c|c|c|}
\hline
\thead{Cultivar}           & \thead{2}     & \thead{5}     & \thead{10}    & \thead{20}    & \thead{All}   \\ \hline
Riesling (MTL)                 & 2.25 & 2.16 & 2.10 & 1.71 & 1.97 \\ \hline
Riesling (STL)           & 4.59 & 4.40 & 3.66 & 3.41 & 3.63 \\ \hline
Cabernet Sauvignon (MTL)       & 2.02 & 2.16 & 2.27 & 2.07 & 2.27 \\ \hline
Cabernet Sauvignon (STL) & 2.68 & 3.24 & 3.68 & 2.91 & 3.43 \\ \hline
Merlot (MTL)                    & 1.69 & 1.54 & 1.55 & 1.41 & 1.53 \\ \hline
Merlot (STL)             & 2.26 & 1.99 & 1.83 & 1.67 & 1.66 \\ \hline
\end{tabular}%
}
\caption{Measuring the impact of varying the dataset size for chosen cultivars. The experiment is conducted for both STL and MTL. We choose 2, 5, 10, and 20 seasons as reasonable choices to evaluate. The performance is measured in terms of RMSE.}
\label{tab:datasetsize}
\end{table}
\begin{table}[t]
\centering \fontsize{9}{9}\selectfont
\resizebox{0.98\columnwidth}{!}{
\begin{tabular}{|l|c|c|c|c|c|}
\hline
\thead{Cultivar}           & \thead{High}  & \thead{Low}   & \thead{Mix}   & \thead{All}   & \thead{Single} \\ \hline
Riesling           & 1.95 &    & 1.70 & 1.97 & 3.63  \\ \hline
Cabernet Sauvignon & 2.49 &    & 2.28 & 2.27 & 3.43  \\ \hline
Chardonnay         & 1.33 &    & 1.41 & 1.40 & 1.60  \\ \hline
Concord            & 2.00 &    & 1.90 & 1.98 & 2.61  \\ \hline
Merlot             & 1.46 &    & 1.39 & 1.53 & 1.66  \\ \hline
Syrah              & 1.13 &    &   & 1.29 & 1.57  \\ \hline
Chenin Blanc       & 1.60 &    &   & 1.45 & 2.47  \\ \hline
Viognier           & 2.60 &    &   & 2.28 & 4.16  \\ \hline
Malbec             & 1.12 &    &   & 0.96 & 1.71  \\ \hline
Pinot Gris         & 1.48 &    &   & 1.61 & 2.04  \\ \hline
Barbera            &   & 2.93 &   & 1.89 & 4.22  \\ \hline
Grenache           &   & 2.04 &   & 1.79 & 2.86  \\ \hline
Nebbiolo           &   & 1.85 &   & 1.24 & 2.48  \\ \hline
Zinfandel          &   & 2.07 &   & 1.60 & 2.64  \\ \hline
Semillon           &   & 2.47 &   & 1.75 & 3.58  \\ \hline
Mourvedre          &   & 1.93 & 1.75 & 1.56 & 2.25  \\ \hline
Sauvignon Blanc    &   & 1.65 & 1.27 & 1.22 & 1.71  \\ \hline
Gewurztraminer     &   & 1.83 & 1.35 & 1.20 & 2.70  \\ \hline
Lemberger          &   & 2.24 & 1.50 & 1.49 & 3.23  \\ \hline
Cabernet Franc     &   & 2.91 & 2.13 & 2.39 & 4.00  \\ \hline
\end{tabular}%
}
\caption{Measuring the impact of choosing a subset of available tasks for MTL and how it fares against choosing all tasks.}
\label{tab:tasksubset}
\end{table}

{\bf Impact of Task Dataset Size.} Table \ref{tab:datasetsize} shows the performance of MTL(MultiHead) and STL models when we choose one of three cultivars (Riesling, Merlot, Cabernet Sauvignon) with $\sim$30 seasons of data and artificially select only a subset of seasons, and train the MultiHead model. We also train STL models in the same setup. As expected for STL, with the exception of Cabernet Sauvignon, introducing more seasons of data up to an extent does help in improving performance.\footnote{Note that there is a consistent decrease in performance when going from 20 seasons to ALL. The reasons for this remain to be explored; however, it is likely due to the influence of a small number of unusual seasons.}  

Interestingly we see that an MTL model trained on just 2 or 5 seasons of data for a cultivar outperforms an STL model using all of that cultivar's data. This reflects the fact that MTL is indeed able to leverage the information present in other cultivars to learn a good model for that specific cultivar. In a sense, the data from other cultivars appear to be as valuable as tens of seasons of data for a specific cultivar. 

%In the case of MTL, adding more data again helps in improving performance but only up to a certain extent. The trend observed in Cabernet Sauvignon is not clearly understood, the performance seems to degrade beyond 2 seasons of data. A closer inspection showed that there was high variance in the performance for the three train/test splits in this case, the exact reason for this variance remains to be explored in future work.

{\bf  Impact of the number of tasks} - The goal here is to understand how different subsets of tasks impact the performance of an MTL model. Here, we select different subsets of our tasks, 10 tasks with the most amount of data, 10 tasks with the least amount of data, and 10 tasks with a mix of high and low amounts of data. We train the MultiHead architecture for this experiment.

Table \ref{tab:tasksubset} presents the cultivars in order of largest to smallest datasets. Each column corresponds to the subset of cultivars used in each experiment. Interestingly, we observe that an MTL model trained on any of our chosen subsets always outperforms single-task models for all cultivars. 

For cultivars with relatively higher amounts of data, surprisingly, it is better to choose a mix of high and low data cultivars to get a better performing model. Including all the cultivars for training does not lead to consistent gains for all cultivars with high amounts of data. The reasons for this observation require further analysis and experimentation.  

For cultivars with lower amounts of data, again, choosing a mix of high and low data cultivars leads to a better performing model. Including all the cultivars for training does indeed lead to consistent gains over choosing a subset. 

{\bf Impact of Training Setting.} Here, we consider a different training setting, Transfer Learning \cite{transferlearning}, where a new cultivar arrives and is incorporated into the model without access to past data. 
%look at Transfer Learning  as an alternative to Multi-Task Learning. 
This is in contrast to MTL where all datasets are accessible at training time. We consider finetuning as a straightforward approach to transfer learning. Finetuning, in the case of MultiHead refers to replacing the task-specific final layers with a newly initialized layer for the new task. For the task embedding approaches, finetuning refers to learning the coefficients of a linear combination of existing task embeddings.  

Table \ref{tab:trainingsetting} shows the RMSE metrics for the finetuning paradigm for the different proposed methods relative to their corresponding RMSE metrics for MTL from Table \ref{tab:mainresults}.

In the case of the MultiHead architecture, we observe that finetuning is on par with MTL. This seems to indicate that there are no tasks that hurt the MTL training process. 

Although, for the Task Embedding approaches, we see that finetuning does worse than MTL for most cultivars for the Concatenate and Additive variants. For the multiplicative embedding variants, we see marginal to substantial gains in performance. 

%\section{Limitations}
%We notice that for cultivars with extremely sparse data, like Cabernet Franc, MTL can't fully alleviate overfitting.
\begin{table}[t]
\centering \fontsize{9}{9}\selectfont
 \resizebox{0.99\columnwidth}{!}{
\begin{tabular}{|l|c|c|c|c|}
\hline
\multicolumn{1}{|c|}{\textbf{Cultivar}}     &\makecell{\textbf{ConcatE} \\ \textbf{FT}} &\makecell{\textbf{MultE} \\ \textbf{FT}} &\makecell{\textbf{AddE} \\ \textbf{FT}} &\makecell{\textbf{MultiH} \\ \textbf{FT}}\\ \hline
Barbera            & -1.02 &  0.04  & -1.69 &  0.01  \\ \hline
Cabernet Franc     & -1.50 &  2.41  & -2.28 & -0.05  \\ \hline
Cabernet Sauvignon & -1.02 &  0.64  & -1.29 & -0.01  \\ \hline
Chardonnay         & -1.14 &  0.04  & -3.01 &  0.11  \\ \hline
Chenin Blanc       & -0.74 &  0.35  & -2.96 & -0.05  \\ \hline
Concord            & -3.38 &  0.11  & -2.96 & -0.24  \\ \hline
Gewurztraminer     & -1.58 &  0.51  & -2.66 & -0.25  \\ \hline
Grenache           & -0.32 &  1.27  & -1.69 & -0.01  \\ \hline
Lemberger          & -2.53 &  1.37  & -0.43 & -0.16  \\ \hline
Malbec             & -2.78 &  0.77  & -1.66 & -0.07  \\ \hline
Merlot             & -0.84 &  0.32  & -2.32 &  0.11  \\ \hline
Mourvedre          & -1.28 &  0.22  & -1.58 & -0.07  \\ \hline
Nebbiolo           & -2.51 &  0.72  & -1.31 & -0.41  \\ \hline
Pinot Gris         & -1.14 &  0.53  & -2.44 &  0.08  \\ \hline
Riesling           & -1.78 &  1.14  & -1.00 &  0.31  \\ \hline
Sangiovese         & -0.96 &  0.31  & -1.73 &  0.05  \\ \hline
Sauvignon Blanc    & -1.22 &  0.09  & -0.23 & -0.02  \\ \hline
Semillon           & -1.46 &  0.97  & -3.08 &  0.35  \\ \hline
Syrah              & -0.97 & -0.06  & -1.87 &  0.00  \\ \hline
Viognier           & -1.25 &  2.13  & -1.89 &  0.52  \\ \hline
Zinfandel          & -2.24 &  1.31  & -1.75 & -0.19  \\ \hline\hline
\textbf{Median}             & -1.25 &  0.53  & -1.75 & -0.01  \\ \hline 
\textbf{Mean}               & -1.51 &  0.72  & -1.90 &  0.00  \\ \hline
\end{tabular}%
 }
\caption{Comparing Transfer Learning with Multi-Task Learning. Note that the performance is relative to corresponding MTL counterparts in table \ref{tab:mainresults}. If a term is positive, it means that transfer learning does better than MTL in that case. We abbreviate finetuning with FT in the column names.}
\label{tab:trainingsetting}
\end{table}
\section*{Path to Deployment}

Farmers use AgWeatherNet \cite{AgWeatherNet} and WSU Viticulture and Enology \cite{viticulture_wsu} websites to monitor cold hardiness through the deployment of the Ferguson model and publication of real LTE values, respectively. Our goal is to finalize the MTL-based models proposed in this paper and deploy them onto AgWeatherNet for the 2022-2023 season for beta testing. %A subset of users will be given access to the models and asked for feedback at the end of the season. Farmers will be able to use weather data from the nearest AgWeatherNet station. We would use ONNX \cite{onnx} to make our models user-friendly, portable, and inference ready.

\section*{Conclusion}

We showed that multi-task learning is an effective approach to predicting Cold Hardiness for grapevines. In particular, our model consistently outperforms the state-of-the-art scientific model without relying on expert domain knowledge. This model will be deployed on an existing weather network for the 2022-2023 season. In the future, we plan to apply these ideas to cold-hardiness prediction for other crops, such as cherries and apples. In addition, we plan to investigate the utility of MTL for other agriculture-related problems with limited data. 

\section*{Acknowledgements}
%This research was supported by NSF and USDA-NIFA under the AI Institute: Agricultural AI for Transforming Workforce and Decision Support (AgAID) award No.2021-67021-35344. The authors thank Lynn Mills and Alan Kawakami for the collection of LTE data.
This research was supported by USDA NIFA award No. 2021-67021-35344 (AgAID AI Institute). 
The authors thank Lynn Mills and Alan Kawakami for the LTE data collection.%the Keller lab at WSU for LTE data collection.

\bibliography{aaai23.bib}
\end{document}